\documentclass[acmlarge]{acmart-ecaf}

\renewcommand{\baselinestretch}{1.15}

\AtBeginDocument{%
  }

\setcopyright{cc}
\copyrightyear{2026}
\acmYear{2026}
\acmConference[ECAF'26]{Fifth European Conference on Algorithmic Fairness}{September 02--September 04, 2026}{Ghent, BE}

\usepackage{multirow}
\usepackage{arydshln}
\usepackage{subcaption}

\begin{document}

\title{WIDER-FAIR: An Annotated Version of the WIDER-FACE Dataset for Fairness Evaluation}


\author{Maxime Moussi}
\affiliation{%
  \institution{UCLouvain}
  \city{Louvain-la-Neuve}
  \country{Belgium}
}

\author{Beno\^it Ronval}
\affiliation{%
  \institution{UCLouvain, ICTEAM}
  \city{Louvain-la-Neuve}
  \country{Belgium}
}
\email{benoit.ronval@uclouvain.be}

\author{Siegfried Nijssen}
\affiliation{%
  \institution{UCLouvain, ICTEAM}
  \city{Louvain-la-Neuve}
  \country{Belgium}}
\affiliation{%
  \institution{KU Leuven, DTAI}
  \city{Leuven}
  \country{Belgium}
}

\author{Félicien Schiltz}
\affiliation{%
  \institution{Euranova}
  \city{Mont-Saint-Guibert}
  \country{Belgium}
  \authornote{At the time of publication, Félicien Schiltz is employed by the European Data Protection Supervisor. The views expressed in this text do not necessarily reflect those of the EDPS}
}

\renewcommand{\shortauthors}{M. Moussi et al.}

\begin{abstract}

The deployment of face detection models in real-world applications raises important fairness concerns, as these systems may showcase performance disparities across demographic groups. A key obstacle to studying and mitigating such biases is the lack of face detection datasets with sensitive feature annotations. To address this gap, we introduce WIDER-FAIR, a new dataset built on the widely used WIDER-FACE benchmark, manually annotated with the perceived ethnicity and sex of each face. The dataset contains 16,256 images annotated across four ethnic groups: Asian, Black, Indian, and White, and two sex categories. We assess the quality and coherence of the annotations using face embeddings, a K-Nearest Neighbors classifier, and a t-SNE visualization, all of which support the consistency of the labeling process. As a demonstration of the dataset's potential, we train a YOLOv5 model and perform ablation studies on each sensitive feature. Among other findings, our experiments show that detection performance is notably lower for faces of Black individuals, and that excluding this group from training increases fairness disparity more than excluding any other ethnic group. These observations illustrate the value of demographically annotated datasets for understanding and evaluating bias in face detection models.

\end{abstract}


\maketitle

\section{Introduction}






As Machine Learning (ML) models are increasingly developed and deployed in high-stakes domains from hiring to medical diagnosis and law enforcement, the concerns about their fairness and potential for discriminatory outcomes have grown substantially. In many regions, legislation is starting to mandate that ML models must ensure fairness toward protected classes to remain compliant. For example, the EU AI Act \cite{EUAIAct2024} explicitly requires systems to detect, prevent, and mitigate risks of bias and discrimination.
Thus, these ML models should produce an outcome that is independent of one or more sensitive features, which are attributes that could be used to distinguish different demographic groups. Examples of sensitive features include the ethnicity, the sex, or the age of a person, among other possible elements. However, without proper data processing, algorithm adaptation, or processing of the outputs, most standard ML models may showcase biased outcomes \cite{mehrabi2021survey}, which must be detected for their safe release and application in the real world.
\\


For these reasons, it becomes more and more important to have access to datasets and to develop evaluation procedures that can be used to assess the potential biases of these models. Such datasets should be usable for standard tasks, such as classification or regression, but also contain information regarding the sensitive features, such that the models used for a given task can be evaluated for fairness regarding the involved demographic groups. Along with the datasets, methods to assess fairness in the considered models should also be explored to detect biases, understand their potential source, and provide hints on how to correct them.
\\

Among the large variety of possible applications of Machine Learning, we focus in this work on the face detection task. The goal of this task is to predict the positions of the faces in a given image. These pictures may be highly diverse, from ID card images to photos of a crowd during an event.
Thus, each image may contain from one to many faces to be detected by a model, in very different conditions regarding the lighting, the poses of the people, the potential partial obstructions of the faces, or even the quality of the photo itself.
This diversity in the images makes the face detection itself a challenging task, which may require dedicated models to achieve good performance. Moreover, this task also presents challenges regarding fairness \cite{mittal2023face}. The face detection model should identify the faces of people regardless of sensitive attributes, such as their ethnicity or their sex. It is therefore important to have access to datasets that contain this diversity of images and that contain information about sensitive features, such that the models used for face detection can be evaluated both on their actual performance and on their discriminative biases.
\\

In this work, we present a new dataset called WIDER-FAIR. It contains a subset of images from the WIDER-FACE dataset \cite{yang2016wider}, which is annotated by hand by a human with information related to ethnicity and sex for each face in the images. We distinguish between four ethnicities: Asian, Black, Indian, and White, and we consider both male and female for the sex. We focus on these two categories of sensitive features as they are among the most used criteria for fairness evaluation \cite{mehrabi2021survey,karkkainen2021fairface,mittal2023face} and are generally protected by anti-discrimination laws. In this paper, we use the ethnicity rather than another similar sensitive feature, such as skin tones as in previous work \cite{Yang2022}, given that legislation such as the EU law generally define the protected features with the ethnic origins. It is important to highlight that whenever we refer to the ethnicity or the sex, we actually refer to the \textbf{perceived} ethnicity or sex, based only on the images from the dataset and the human annotator's judgment, as the original dataset does not contain any of such information. Thus, in the cases where the values for at least one of these sensitive features is unsure, the whole image is ignored to avoid forcing an uncertain label on a face, which would introduce a potential bias in the dataset.

The quality of the annotations is then investigated with different methods directly inspired by the literature \cite{karkkainen2021fairface}. A classifier is learned to predict the sensitive features and its low error rate hints to the correctness of the dataset, and a data visualization method is used to represent the data and ensure that faces with the same sensitive values cluster together.

In a second part, we illustrate a potential use of the annotated dataset by studying the impact of each group of the two sensitive features on the performance of the face detection task. This investigation is done by performing ablation studies of each group, one at a time. This approach is performed to identify potential causes of bias that could appear when training a face detection model on our annotated dataset without proper fairness-aware approaches.

We believe this dataset can be interesting to the community as it builds on a well-known dataset for face detection and contains information that can be useful for fairness evaluation in such tasks. The dataset, along with the code for the experiments in this paper, is available in the following repository: \url{https://github.com/bronval/Wider-Fair-Dataset}.
We insist that all the annotations in this dataset were carried out by a single person. While this process was done in good faith and with the explicit goal of supporting fairness evaluation, the dataset can not be guaranteed to be free of unintended bias or annotation errors. As such, future researchers are encouraged to treat this dataset with the appropriate caution.

\section{Background and Related Work}





\subsection{Face Detection Task}

Face detection is the task of automatically locating human faces within an image, typically by predicting the position of a \textit{bounding box} around each face present in the scene. A bounding box is a rectangular region defined by a set of coordinates, along with its height and width, that tightly encloses a detected face. It is important to distinguish face detection, studied in this work, from the related but distinct tasks of face recognition and face classification. Face recognition aims to identify who a person is by matching a detected face against a database of known individuals, while face classification assigns a label to a detected face \cite{zafeiriou2015survey}. Face detection is thus a prerequisite for both of these tasks, but is only concerned with the positions of the faces in an image, without any attempt to predict their identity or attributes. 

Different models have been designed to perform face detection. Among others, MTCNN \cite{zhang2016joint} and RetinaFace \cite{deng2020retinaface} are deep models that use convolutional neural networks to detect faces in images. RetinaFace goes a step further by reconstructing the mesh of the detected faces.

\subsection{YOLO Model for Object and Face Detection}

YOLO (You Only Look Once) \cite{redmon2016you} is another family of deep models used for object and face detection.
YOLO divides the input image into a grid of cells, each of which simultaneously predicts a set of bounding boxes and associated class probabilities in a single forward pass through the network. This architecture makes YOLO substantially faster than two-stage detectors while maintaining competitive accuracy, enabling real-time detection on standard hardware.
The different versions of YOLO models have been widely used in various image tasks \cite{terven2023comprehensive,diwan2023object}. More specifically, prior research on YOLOv5 has shown that minor modifications can lead to strong results for the face detection task \cite{qi2022yolo5face}. Thus, in this work, we use the YOLOv5 model to assess the impact of the different sensitive groups on the performance of this model.

\subsection{Datasets for Face Detection}

Different datasets have been proposed in the literature regarding the face detection task. In this subsection, we focus on three of them. 
The WIDER FACE dataset \cite{yang2016wider} is one of the most widely used benchmarks for face detection. It contains 32,203 images with 393,703 faces, where each face is annotated with its corresponding bounding box in its corresponding image. Each face is further annotated with the occlusion level, pose, and event category, ranging from parades and concerts to sports events and traffic. This diversity in the events and the wide range of difficulty due to the different occlusion levels, image scales, or lighting make this dataset a particularly relevant benchmark for the face detection task.
However, the WIDER-FACE dataset contains no demographic annotation. This lack of information regarding the ethnicity, sex, or age of the people's faces strongly limits the use of this dataset to assess the fairness of a model.
\\

The limitation regarding the demographic annotation has been addressed by a more recent related work \cite{Yang2022}, in which the WIDER-FACE dataset is annotated with the sex, age, and skin tones. These values are obtained through the use of a commercial annotation service with human evaluators. Moreover, a face attribute prediction model based on FairFace \cite{karkkainen2021fairface} is used to stop sampling images of faces that are overly represented in the dataset. Overall, this dataset contains more than 35,000 annotated faces.

This prior work is the most similar to the dataset we propose. However, we identify key differences that we believe make our work a valuable contribution. First, this prior work does not consider ethnicity labels and rather uses skin tones. Although the choice between the two setups can be discussed, we believe that the inherent difficulty of the WIDER-FACE dataset, especially with the different lighting of the images, makes the annotations of the skin tones much more difficult than the prediction of the ethnicity. Moreover, it is worth noting that in legislation such as EU law, protected features are defined as racial or ethnic origin rather than skin tone, which aligns more closely with our work.
The second difference is the use of a model to automatically stop sampling images from certain sensitive groups. This may introduce a bias due to the classification model itself, its learning algorithm, or its training data. This may lead to biased data, where the bias could then be transferred to a later model using the data as a training or test set.
\\

Another image dataset is FairFace \cite{karkkainen2021fairface}, which differs from the others as it uses the images from the YFCC-100M Flickr dataset \cite{thomee2016yfcc100m}. It contains 108,501 images with annotations regarding the sex, the age, and the ethnicity, for which it considers seven distinct values: White, Black, Indian, East Asian, Southeast Asian, Middle Eastern, and Latino. However, this dataset is primarily designed for face attribute classification rather than face detection. This is illustrated by the images being tightly cropped, single face, and without any annotation on bounding boxes. These design choices in the FairFace dataset prevent its use to train and assess face detection models.

\subsection{Fairness in Face Detection}

Regarding the specific impact of sensitive groups on the performance of models, different prior works present conclusions that may sometimes vary. Concerning the classification itself, dark-skinned individuals and women tend to be disadvantaged, i.e., less often detected, regardless of the balancing methods of the datasets \cite{karkkainen2021fairface,buolamwini2018gender}.
On the specific face detection problem, a first work \cite{mittal2023face} reports that female individuals appear to have an advantage over males in terms of detection performance, while another work \cite{menezes2021bias} reports the opposite conclusion. In all cases, dark-skinned people tend to be disadvantaged for this task compared to other possible sensitive subgroups \cite{mittal2023face,wilson2019predictive,Yang2022,menezes2021bias}.

\section{WIDER-FAIR Dataset}




In this section, we first detail the process that was used to annotate the WIDER-FACE dataset. Then, we explore the resulting dataset using different statistics and representations to ensure that the global aspect of the dataset is valid.

\subsection{Annotation Process}

\subsubsection{WIDER-FACE Choice}

We built the dataset using the images from WIDER-FACE. We select this dataset as it is one of the most used benchmarks for face detection, and because it contains a wide variety of images, as explained in the previous section. We believe that building on top of this commonly used benchmark can allow for easy fairness assessment of most existing models for face detection.

\begin{table}[t]
\centering
\caption{Repartition of the images regarding the ethnicity and sex in the annotated dataset}
\begin{tabular}{c|cc|cc|cc|cc}
 & \multicolumn{2}{c|}{\textbf{Asian}} & \multicolumn{2}{c|}{\textbf{Black}} & \multicolumn{2}{c|}{\textbf{Indian}} & \multicolumn{2}{c}{\textbf{White}} \\
 & Male & Female & Male & Female & Male & Female & Male & Female \\
\hline
Counts      & 779  & 899  & 1034 & 609  & 769  & 638  & 6211  & 5317  \\
Proportion (\%)& 4.79 & 5.53 & 6.36 & 3.75 & 4.73 & 3.92 & 38.21 & 32.71 \\
\end{tabular}
\label{tab:dataset}
\end{table}

\subsubsection{Annotation Pipeline}
\label{sec:pipeline}

\begin{figure}[t]
    \centering
    \includegraphics[width=0.95\linewidth]{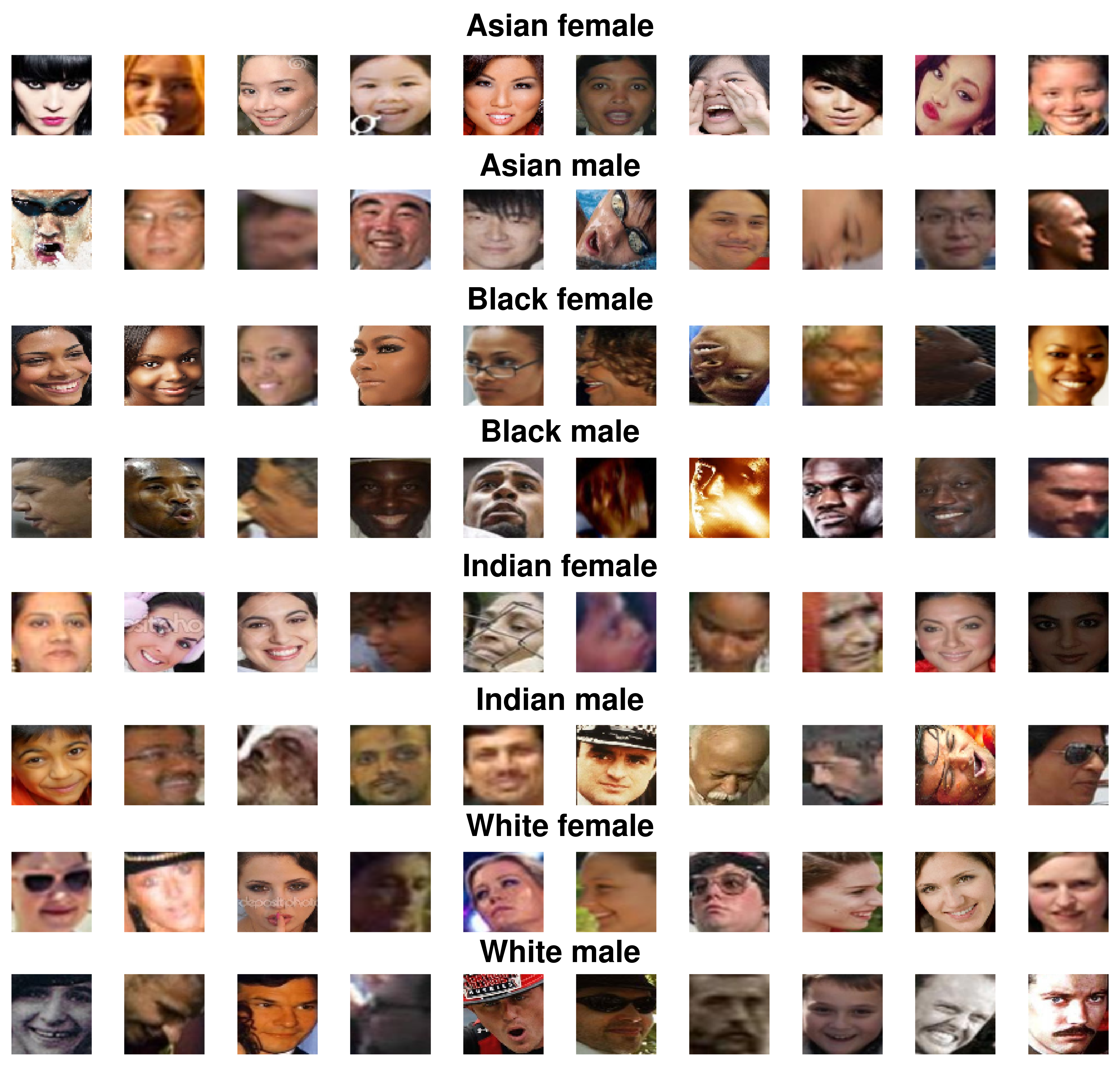}
    \caption{Samples from the WIDER-FAIR dataset}
    \label{fig:data_samples}
\end{figure}

The first step of the annotation pipeline is to remove images that contain faces that may be too difficult to annotate and later detect. Indeed, as the WIDER-FACE dataset contains many images with varying quality, some of these images include very small faces, faces that are almost entirely hidden (occlusion), or very poor image quality due to blurry effects. These factors may impact the performance of a prediction model, but, more importantly, may alter the conclusion of the observation regarding the impact of the sensitive groups on the performance of the model, as we do not know the distribution of the image difficulty for each sensitive group. For example, if most images with Asian people are more blurry than the others, the results for this sensitive group may be impacted negatively for a reason that is independent of the sensitive feature.

Therefore, we remove all images that match at least one of the three following criteria:
\begin{itemize}
    \item \textbf{Extreme occlusion}, corresponding to bounding boxes annotated with the highest level of occlusion.
    \item \textbf{Extreme blur}, corresponding to bounding boxes annotated with the highest level of blur.
    \item \textbf{Extremely small bounding boxes}, corresponding to bounding boxes with a relative area less than $0.15\%$ of the image.
\end{itemize}
To be consistent, we remove an image if at least one of its bounding boxes matches one of the criteria.
\\

The second step is the annotation for the sensitive features. We consider two features: the ethnicity and the sex of the people in the images. For the ethnicity, we distinguish between four values: Asian, Black, Indian, and White. Although more or different values could be considered, we find these four categories to cover a large set of the images from WIDER-FACE, while keeping the analysis and fairness evaluation both valuable and concise.
\\

A human annotator was then tasked to review the images from the WIDER-FACE dataset and complete the \textbf{perceived} values for the sensitive groups listed above. During this step, the annotator had the possibility to encode two additional values for the sensitive groups: \textbf{Undetermined}, whenever the face was too difficult to classify due to the image quality or simply because the sensitive value was unclear, and \textbf{Other}, when the ethnicity of the presented face did not match any of the four values listed above, for example for Native American. These two values are introduced to allow the annotator to skip an image if the decision is not certain enough, preventing a forced choice for one of the predetermined categories. The whole image was removed if it contained at least one unidentified face to avoid non-annotated faces in the training and test sets.
\\

Although we refer to these features as ethnicity and sex, it is again important to highlight that the reported values correspond to the \textbf{perceived} ethnicity and sex, and that errors are possible, but in no case intentional or motivated with bad intentions. The whole annotation process was conducted in good faith, with the hope that the created dataset can be used to assess and improve fairness in models used for face detection.

\begin{figure}[t]
\centering
\begin{minipage}{0.48\linewidth}
    \centering
    \includegraphics[width=1.0\linewidth]{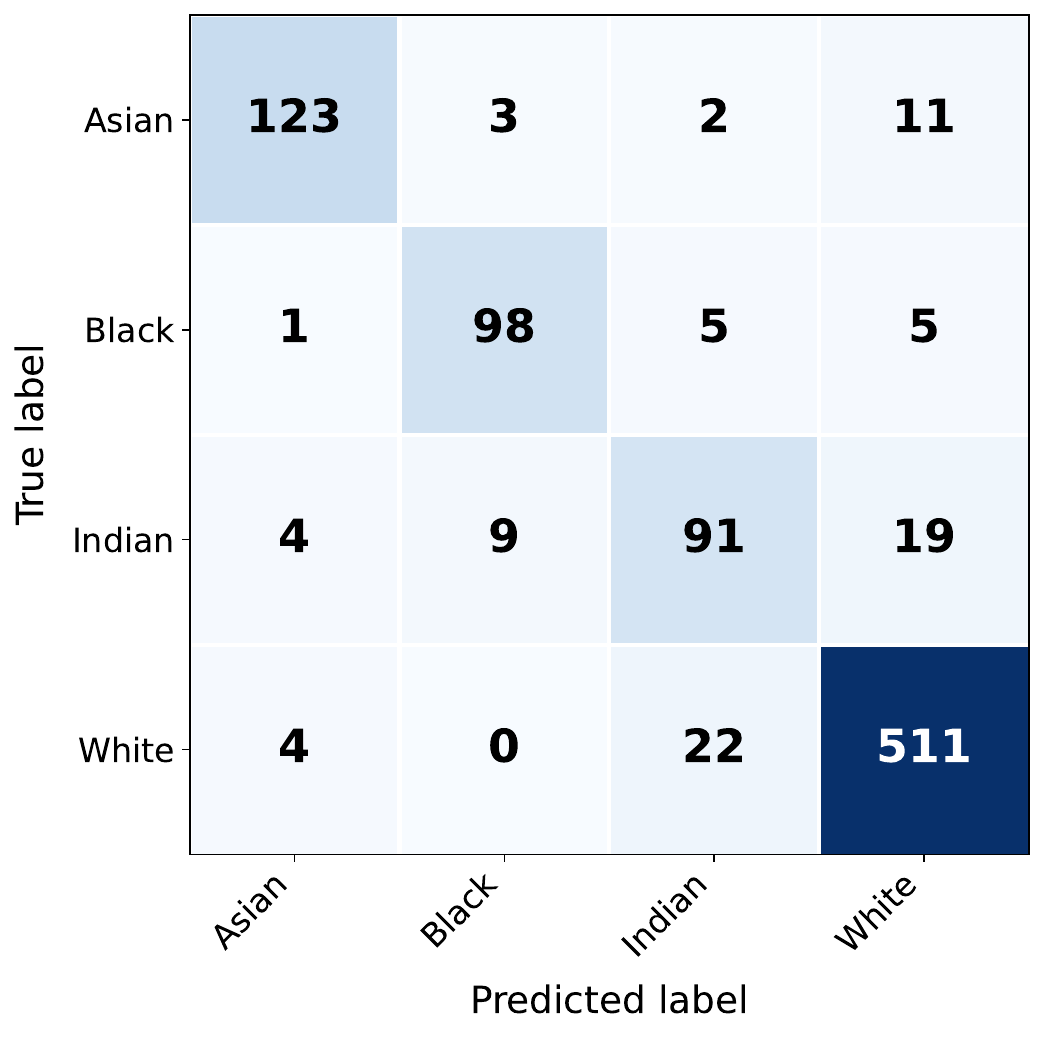}
    \caption{Confusion matrix for ethnicity classification}
    \label{fig:conf_ethn}
\end{minipage}%
\begin{minipage}{0.48\linewidth}
    \centering
    \includegraphics[width=1.0\linewidth]{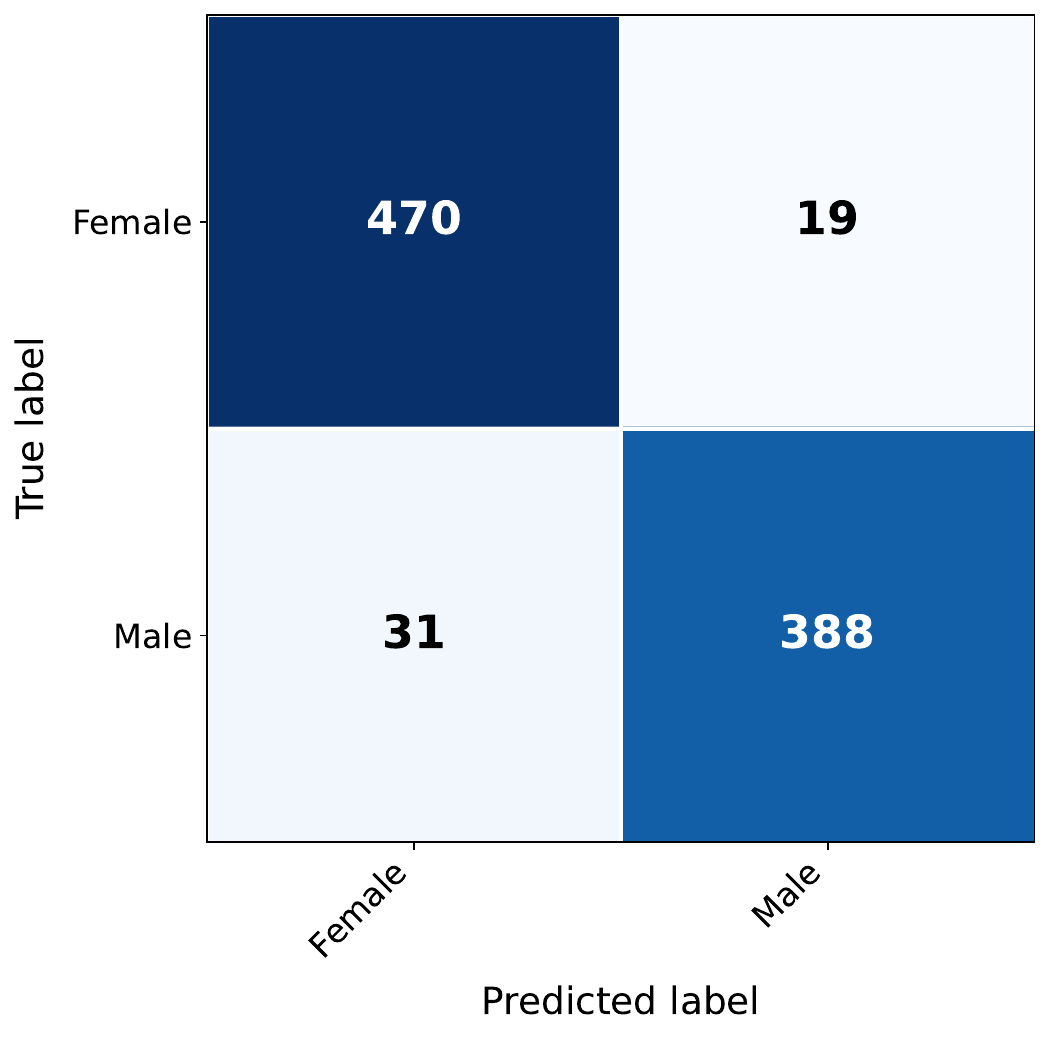}
    \caption{Confusion matrix for sex classification}
    \label{fig:conf_sex}
\end{minipage}
\end{figure}

\begin{figure}[t]
    \centering
    \includegraphics[width=0.9\linewidth]{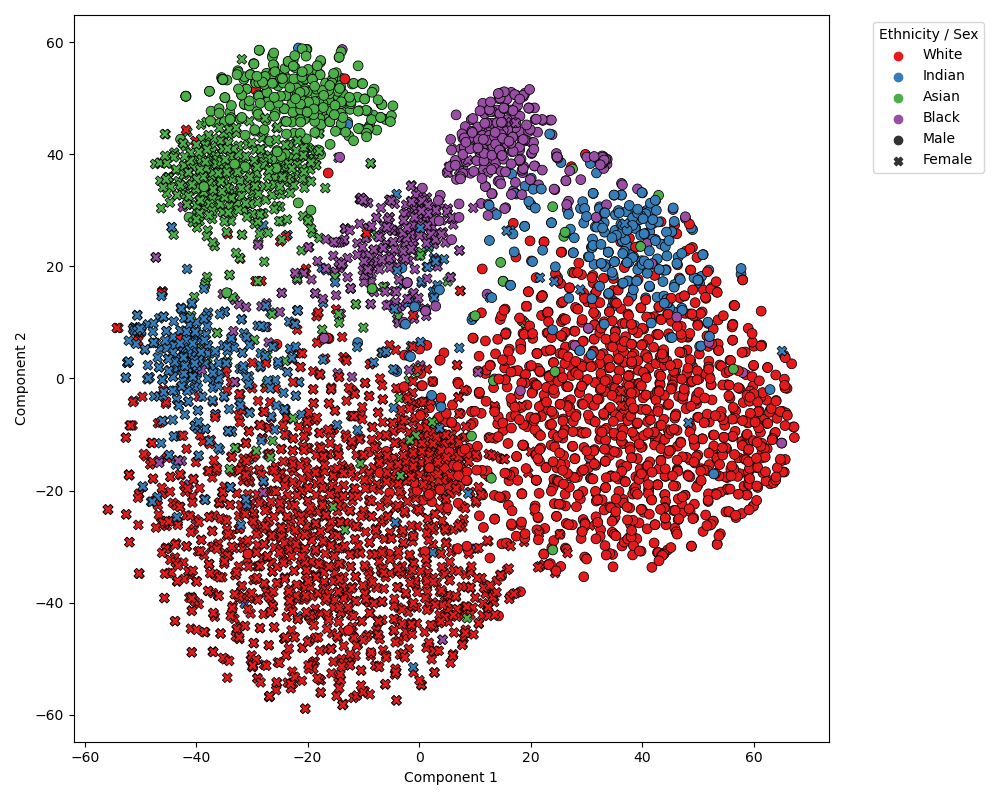}
    \caption{t-SNE representation of the annotated images}
    \label{fig:tsne}
\end{figure}

\subsection{Dataset Analysis}

\subsubsection{Sensitive Features Distribution}
At the end of the process, we have annotated a total of 16,256 images, representing more than half of the images from the WIDER-FACE dataset. Our dataset contains $45\%$ of females against $55\%$ of males, a limited imbalance compared to datasets used in other works \cite{panagiotou2025tabfairgdt,mittal2023face}. However, as the base dataset is imbalanced towards the representation of White people, our annotated dataset reflects this imbalanced distribution. The complete repartition of the sensitive groups is detailed in Table \ref{tab:dataset}.

\subsubsection{Dataset Examples}

Figure \ref{fig:data_samples} displays examples from the annotated dataset. These examples show that there is an important intra-class variability, due to both the individual traits of the people, as well as the image quality inherent to the WIDER-FACE dataset. A particular example of this is the two pictures of Barack Obama in the Black male group, more precisely, the first and the third pictures, where he appears with different skin tones due to the lighting. This reinforces us in the choice of labeling the images with the ethnicity instead of the skin tone, as done in other works \cite{Yang2022}.

\subsubsection{Classification and 2D Representation}

To investigate further the quality of the annotation, and following the procedure used in FairFace, we study the distribution of the data in an embedding space. In this setup, only the simplest images are considered, i.e., the images with the lowest levels of occlusion or blur, to ensure that the embedding model is not influenced by external factors. A face recognition model \cite{king2009dlib} is used to obtain embedding vectors of size 128 to represent the faces.

First, a K-Nearest Neighbors (KNN) classifier is applied on these vectors, with the goal to predict the sensitive values of the given faces. The resulting confusion matrices are displayed in Figure \ref{fig:conf_ethn} for the ethnicity and in Figure \ref{fig:conf_sex} for the sex. In both cases, the large majority of images are correctly classified, which hints that the annotations are at least coherent across images.

A second method to verify the coherence of the annotations is to use the embeddings in a t-SNE representation \cite{van2008visualizing}, ensuring that similar image embeddings are close to each other. As shown in Figure \ref{fig:tsne}, images from the same sensitive groups cluster together, hinting once again that the annotations appear to be coherent.

\section{Evaluation Setup}

\subsection{Goal of the Evaluation}

In the remaining parts of this paper, we aim to study the impact of the presence or absence of the various sensitive groups in the training dataset of a face detection model on its performance and potential bias. We perform ablation studies, each time by removing one of the sensitive groups from the training set and observing the impact on the other groups in terms of detection performance. We consider the YOLOv5 model for our experiments, as it has been shown to obtain strong performance for this type of task \cite{qi2022yolo5face}.

\subsection{Evaluation Metrics}

Regarding the evaluation of the face detection task itself, an important score used in the prediction is the \textit{Intersection over Union} (IoU). It represents how much the area of a predicted bounding box $\hat{b}$ overlaps the area of a ground truth box $b$, and is computed with the following formula
\begin{equation}
    IoU(\hat{b}, b) = \frac{\hat{b} \cap b}{\hat{b} \cup b}
\end{equation}
This score ranges from 0 (no overlap) to 1 (perfect alignment). In practice, a threshold $\tau$ is used to determine if a predicted bounding box matches a ground truth. Thus, a predicted box is considered a true positive (TP) if there is a ground truth box such that $IoU(\hat{b}, b) \geq \tau$. A false negative (FN) corresponds to the case where no ground truth matches the predicted box according to the threshold $\tau$, and a false positive (FP) corresponds to the case where no ground truth box exists.
From these values, the recall for a given threshold $\tau$ is computed to measure the proportion of ground truths detected, with $n$ the total number of faces in the dataset:
\begin{equation}
    Recall(\tau) = \frac{TP(\tau)}{TP(\tau) + FN(\tau)} = \frac{TP(\tau)}{n}
\end{equation}

Regarding the fairness evaluation, we consider the Equal Opportunity, following previous works \cite{mittal2023face,Yang2022}, and also because it focuses on the recall metric. First, we define the recall per group $Recall_g(\tau) = TP(\tau) / n_g$, with $n_g$ being the total number of faces for the sensitive group $g$. From this score, we evaluate the equal opportunity based on the disparity metric, as done in \cite{mittal2023face,wilson2019predictive,Yang2022}. This metric evaluates the standard deviation of the recall between the different sensitive groups for a given threshold $\tau$, thus providing an estimate of the discrepancy in the outputs for the different groups. The disparity is computed with the following formula:
\begin{equation}
    Disparity(\tau) = \sqrt{\frac{1}{K} \sum_{g \in G} (Recall_g(\tau) - \bar{R}(\tau) )^2}
\end{equation}
where $G$ is the set of sensitive groups, $K=|G|$ the number of groups, and $\bar{R}(\tau)$ the mean recall across all groups for the given IoU threshold.

\subsection{Train and Test Splits}

When creating the training and test splits, we follow different constraints to ensure that the experiments are performed in a fair setup. The first constraint is that, if a bounding box of an image is selected in a split, then all the other bounding boxes from that image are included in the same split. Furthermore, in the ablation experiments, we ensure that all the subsamples of the dataset contain the same number of images to ensure that a model is not trained on more data than another model, which could affect the performance and thus the conclusions on fairness, following a setup proposed in a prior work \cite{ronval2024can}.
In all cases, the test set is the same for all the experiments, following the same constraints as above and containing images for all the sensitive groups, regardless of the held-out group. Moreover, the difficulty of the images in the test set is controlled to ensure that each sensitive group has the same distribution of non-demographic factors.

The model is trained three times to account for stochasticity. All the results presented in the next section represent the computed means of these independent runs.

\section{Experiments and Results}

\begin{figure}[t]
\centering
\begin{minipage}{0.45\linewidth}
    \centering
    \includegraphics[width=1.0\linewidth]{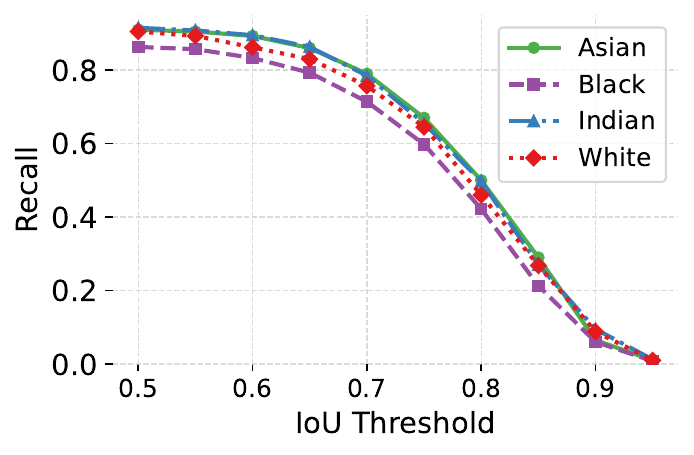}
    \caption{Performance of the base model with the ethnicity sensitive feature}
    \label{fig:base_ethn}
\end{minipage}%
\begin{minipage}{0.45\linewidth}
    \centering
    \includegraphics[width=1.0\linewidth]{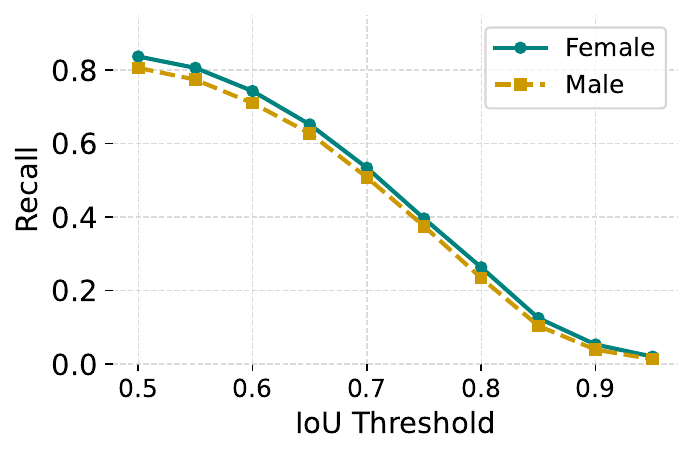}
    \caption{Performance of the base model with sex sensitive feature}
    \label{fig:base_sex}
\end{minipage}
\end{figure}

\subsection{Result on Full Dataset}

The first experiment is to train the YOLOv5 model on that dataset that contains all the ethnicities and sexes to assess the base performance. The results are shown in Figure \ref{fig:base_ethn} for the ethnicity and in Figure \ref{fig:base_sex} for the sex.

In the first case, when varying the IoU threshold $\tau$, we observe there is a clear difference in performance, computed as the recall, for the detection of faces of Black people. The faces of White people appear to be slightly disadvantaged compared to the Indian and Asian groups, which both perform almost equally, regardless of the IoU threshold.
The higher the threshold becomes, the more difficult the predictions become, which is why the scores collapse to a very low recall.

Regarding the second sensitive feature, the gap between the performance of both groups appears to be smaller than in the case of ethnicity, with the male faces being slightly more difficult to detect than the female faces.

As the models are trained on approximately the same number of images for each group in both experiments, with more representation of the White people's faces due to the original dataset's imbalance, and tested on images with similar difficulty, the observed gaps may come directly from the learning process of the model.


\begin{figure}[t]
\begin{minipage}{0.32\linewidth}
    \centering
    \includegraphics[width=1.0\linewidth]{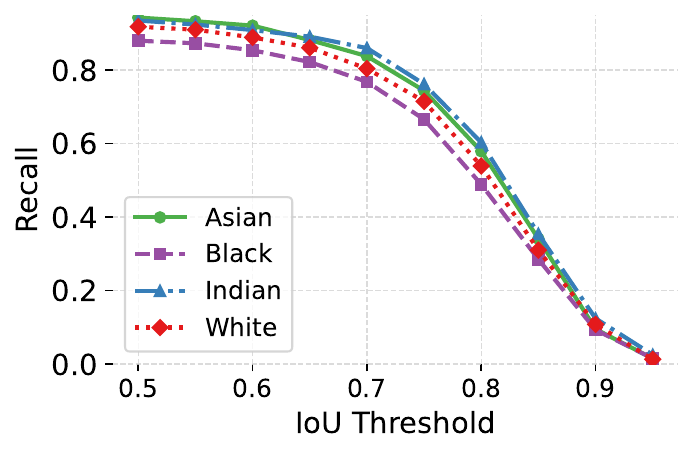}
    \caption{Performance in LOEO Asian}
    \label{fig:loeo_asian}
\end{minipage}%
\begin{minipage}{0.32\linewidth}
    \centering
    \includegraphics[width=1.0\linewidth]{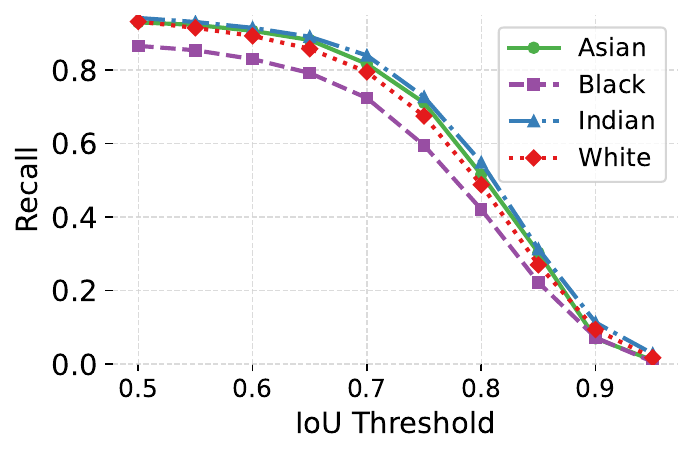}
    \caption{Performance in LOEO Black}
    \label{fig:loeo_black}
\end{minipage}%
\begin{minipage}{0.32\linewidth}
    \centering
    \includegraphics[width=1.0\linewidth]{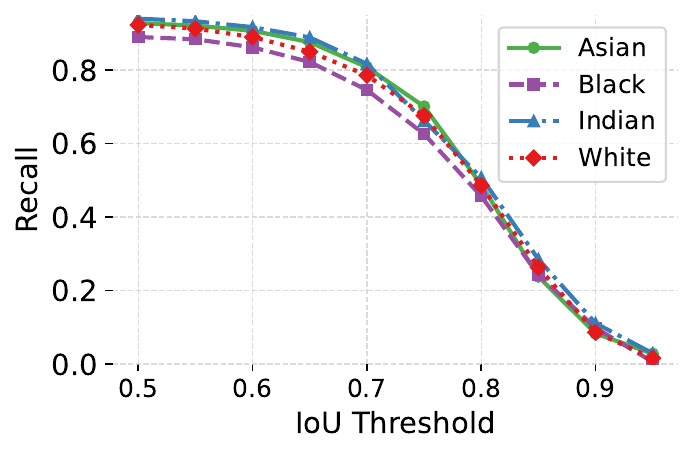}
    \caption{Performance in LOEO Indian}
    \label{fig:loeo_indian}
\end{minipage}
\end{figure}

\begin{table}[t]
\centering
\caption{Detailed results for the recall and the disparity in the Leave One Ethnicity Out experiment, averaged from the IoU thresholds $[0.5, 0.65]$}
\label{tab:loeo}
\begin{tabular}{l|cccc|c}
& \textbf{Asian} & \textbf{Black} & \textbf{Indian} & \textbf{White} & \textbf{Disparity} \\ 
\hline 
Global & 0.892 & 0.837 & 0.895 & 0.873 & 0.024 \\ 
\hdashline
LOEO\_Asian & 0.920 & 0.857 & 0.915 & 0.895 & 0.025 \\ 
LOEO\_Black & 0.911 & 0.835 & 0.920 & 0.900 & 0.033 \\ 
LOEO\_Indian & 0.908 & 0.865 & 0.920 & 0.894 & 0.021 \\ 
\end{tabular}
\end{table}

\subsection{Ablation Studies on Ethnicity}

This next experiment studies the impact of removing a sensitive group of a given ethnicity. We measure this impact by looking at the face detection performance of the other ethnic groups. We denote this experiment by \textit{Leave One Ethnicity Out} (LOEO). For example, LOEO Asian represents the case where the Asian group is removed from the training set. As a reminder, the test set is left unchanged for all experiments.

Figures \ref{fig:loeo_asian}, \ref{fig:loeo_black}, and \ref{fig:loeo_indian} show the results when removing the Asian, Black, and Indian groups. More detailed results are shown in Table \ref{tab:loeo}, with values averaged with the first IoU thresholds, i.e., from 0.5 to 0.65, as this value should not be too low to avoid misclassifying false predictions as positive nor too high to avoid missing good predictions.

Globally, removing one sensitive group has a limited impact on the face detection performance of the other groups, but also on the group itself, compared to the global performance. However, we note that the case where we remove the faces of Black people negatively impacts the prediction performance for this sensitive group when compared to the results of other LOEO models. Although this is expected, as the model did not see any similar image during its training, this is the only scenario where a decrease in performance is observed. Moreover, this scenario increases the disparity more importantly than in the other cases. These observations suggest that the presence of faces of Black people in the dataset is important for the model to generalize correctly.
\\

In conclusion to this experiment, the faces of Black people are still less detected than those of others, but removing a specific ethnic group does not seem to strongly affect the performance of the model, with the notable exception of the faces of Black people.

\begin{figure}[t]
\begin{minipage}{0.48\linewidth}
    \centering
    \includegraphics[width=1.0\linewidth]{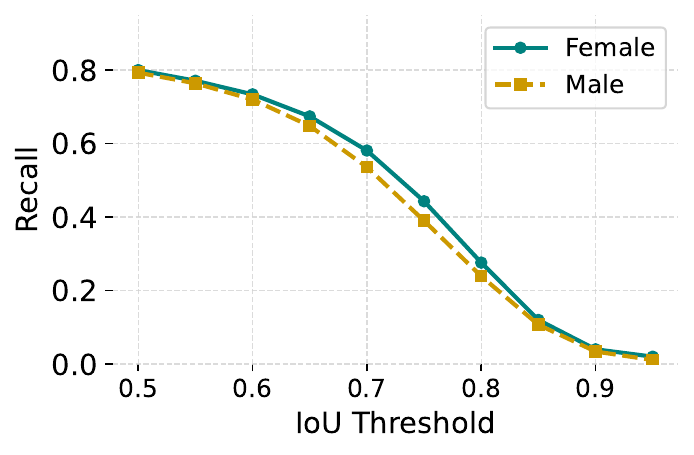}
    \caption{Performance in LOSO female}
    \label{fig:loso_female}
\end{minipage}%
\begin{minipage}{0.48\linewidth}
    \centering
    \includegraphics[width=1.0\linewidth]{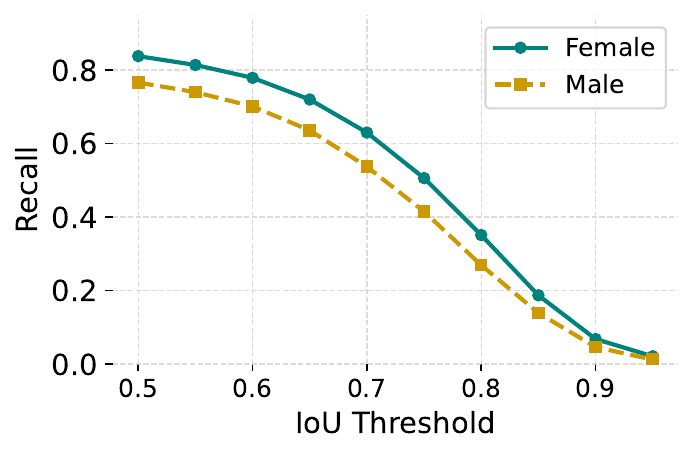}
    \caption{Performance in LOSO male}
    \label{fig:loso_male}
\end{minipage}
\end{figure}

\begin{table}[t]
    \centering
    \caption{Detailed results for the recall and the disparity in the Leave One Sex Out experiment, averaged from the IoU threshold $[0.5, 0.65]$}
\begin{tabular}{l|cc|c}
& \textbf{Female} & \textbf{Male} & \textbf{Disparity} \\ 
\hline 
Global & 0.760 & 0.730 & 0.015 \\
\hdashline
LOSO\_Female & 0.745 & 0.731 & 0.007 \\ 
LOSO\_Male & 0.788 & 0.711 & 0.038 \\ 
\end{tabular}
\label{tab:loso}
\end{table}

\subsection{Ablation Study on Sex}

The impact of sex on performance is studied in a similar setup to the previous experiment. The results with varying IoU thresholds are shown in Figures \ref{fig:loso_female} and \ref{fig:loso_male}, with more detailed results in Table \ref{tab:loso}. We refer to this setup as the Leave One Sex Out (LOSO).

Removing the faces of women reduces the gap between the two groups, but actually reduces the capability of the model to detect female faces. On the other hand, this setup seems to strongly reduce the disparity.

The opposite setup, where male faces are removed instead, increases the disparity, which may hint that the images containing male faces are important to the generalization capabilities of the model.
\\

In conclusion to this experiment, it appears that removing the faces based on the perceived sex leads to results that are more expected than in the LOEO setup, as the prediction performance decreases for the removed group. However, these differences regarding the disparity and the performance hint at the need to have both groups well represented in the data.

\section{Discussion}


As stated before, the proposed dataset has been annotated by hand, by one human, and based on the perceived sensitive values. Although the whole process has been done in good faith with the objective of constructing a dataset that can be used to assess the fairness of face detection models, the annotated dataset may contain unintended errors or biases.
\\

Another element in the annotation process is the level of difficulty. As mentioned in Section \ref{sec:pipeline}, we set a certain threshold for each non-demographic value available in the original dataset above which the image is considered too difficult, but these thresholds are arbitrary and depend on the original annotation of the dataset. In our work, we aim to limit the removal of images based on this criterion by only deleting the most difficult cases, i.e., the most blurry and the most occluded faces, but we recognize this is an arbitrary choice.


\section{Conclusion}


In this work, we proposed the WIDER-FAIR dataset which contains annotations that can be used for most fairness evaluation tasks. This dataset contains images from the widely used WIDER-FACE benchmark, providing annotations on two important sensitive features, the ethnicity and the sex.
As stated in several places of this paper, we insist that the annotations are based on the judgment of one human annotator and based on the \textbf{perceived} ethnicities and sex.
With the annotations done, we explored the coherence of the created dataset by considering a simple classifier, which produces limited mistakes despite the inherent difficulty of the original dataset, and a t-SNE representation where each group appears to form a different cluster, hinting that the annotation are coherent across images. We further demonstrated a potential usage of the dataset to explore and assess the fairness of face detection models by performing ablation studies on each of the sensitive groups.

The dataset will be made public upon acceptance of this paper to encourage future research to ensure that face detection models can be correctly assessed regarding the fairness, an important factor when deploying models in the real world.

\bibliographystyle{ACM-Reference-Format}
\bibliography{sample-base}

\end{document}